# A New Vein Pattern-based Verification System

Mohit Soni
DFS,
New Delhi, INDIA
myk.soni@gmail.com

Sandesh Gupta
UIET, CSJMU,
Kanpur, UP, INDIA
Sandesh@iitk.ac.in

M.S. Rao
DFS,
New Delhi, INDIA
msrnd@rediffmail.com

Phalguni Gupta
IIT Kanpur,
Kanpur, UP, INDIA
pg@iitk.ac.in

*Abstract*— **This paper presents an efficient human recognition system based on vein pattern from the palma dorsa. A new absorption based technique has been proposed to collect good quality images with the help of a low cost camera and light source. The system automatically detects the region of interest from the image and does the necessary preprocessing to extract features. A Euclidean Distance based matching technique has been used for making the decision. It has been tested on a data set of 1750 image samples collected from 341 individuals. The accuracy of the verification system is found to be 99.26% with false rejection rate (FRR) of 0.03%.**

## I. INTRODUCTION

Vein pattern of the palma dorsa can be defined as a random 'mesh' of blood carrying tubes. The back of the hand veins are not deeply placed and hence these can be made visible with the help of a good image acquisition system and technique. The geometry of these veins is found to be unique and universal [14]. Hence, it can be considered as one of the good human recognition systems.

Forensic scientists have always been the biggest reapers of successful biometric systems. User authentication, identity establishment, access control and personal verification etc are a few avenues where forensic scientists employ biometrics. Over time various biometric traits have been used for the above mentioned purposes. Some of them have gained and lost relevance in the course of time. Therefore, constant evolution of existing traits and acceptance of new biometric systems is inevitable. The existing biometric traits, with varying capabilities, have proven successful over the years. Traits like Face, Ear, Iris, Fingerprints, Signatures etc., have dominated the world of biometrics over the years. But each of these biometric traits has its shortcomings. Ear and iris pose a problem during sample collection. Not only is an expensive and highly attended system required for iris but it also has a high failure to enroll rate. In case of ear data, it is hard to capture a non occluded image in real time environment. In case of the most well known face recognition systems there exist some limitations like aging, background, etc [2]. Fingerprints, though more reliable, still lack automation and viability as they are also susceptible to wear and aging. Signatures, are liable to forgery.

Venal patterns, on the other hand, have the potential to surpass most such problems. Apart from the size of the pattern, the basic geometry always stays the same. Unlike fingerprints, veins are located underneath the skin surface and are not prone to external manipulations. Vein patterns are also almost impossible to replicate because they lie under the skin surface [6].

It seems, the first known work in the field of venal pattern has been found in [10]. Badawi [1] has also tried to establish the uniqueness of vein patterns using the patterns from the palma dorsa. The data acquisition technique mentioned in [1] is based on a clenched fist holding a handle to fixate the hand during image capture. This method however, has limitations with respect to orientation. Substantial works in this field have been done by Leedham and Wang [11] [12] [13]. In these, thermal imaging of the complete non fisted hand has been done using Infrared light sources. Generally, near infra-red lamps of intensity-value ranging from 700 to 900 nm in wavelength are used to design such a system [12]. These lamps are found to be costly. Also infra-red light has been used to either reflect or transmit light to the desired surface [8] [11] [12] [14]. These techniques have both their advantages and disadvantages. It has been observed that the images captured through a reflection based system, as proposed in [11], would never produce consistent results owing to excessive noise generated due to unnecessary surface information. The surroundings have to be controlled at all times and the skin color or skin abnormalities are bound to have an effect. The best results can only be expected after exposure from the near infra-red lamps which are costly. A system of capturing images from the front of the hand has been proposed in [14]. The palm prints may interfere with the pattern of the veins, in this case.

Matching technique in a biometric system is a crucial step because the accuracy of the system alone can determine its effectiveness. There exist various matching techniques for proving the individuality of a source. Badawi has used a correlation based matching algorithm and achieved excellent results. However, correlation-based techniques, though the most popular, become costly on larger databases. Wang and Leedham used a matching technique based on Hausdorff distancing, which is limited in principle by the slightest change in orientation.

This paper proposes an efficient absorption based technique for human identification through venal patterns. It makes use of a low cost sensor to acquire images. It uses a fully automated foreground segmentation technique based on active contouring. Reduced manual interference and an automatic segmentation technique guarantee uniform segmentation of all samples, irrespective of their size and position. The paper also employs a rotation and translation invariant matching technique. It is also realized that since the collected images are very large in size (owing to a high definition camera) slow techniques like correlation-based matching would hinder the overall efficiency. Therefore, the proposed system uses critical features of veins to make the decision. The results, thus far, have been found to be encouraging and fulfilling.

Section 2 presents the experimental setup used to acquire images. Next section deals with the proposed venal pattern based biometric system. It tries to handle some of the critical issues such as use of a low cost sensor for acquiring images, automatic detection of region of interest, rotation and translation invariant matching etc. This system has been tested on the IITK database consisting of 1750 image samples collected from 341 subjects in a controlled environment, over the period of a month. Experimental results have been analyzed in Section 3. Concluding remarks are given in the last section.

## II. PROPOSED SYSTEM

Like any other biometric system, the venal pattern based system consists of three major tasks and they are (i) image acquisition (ii) preprocessing of acquired image data (iii) feature extraction (iv) matching. The flow diagram of the proposed system is given in Fig. 1.

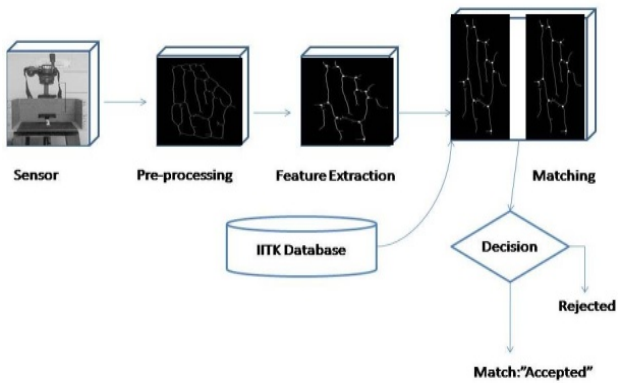

Figure 1. Flow Diagram of the Proposed System

### A. Image Acquisition

Performance of this type of system always depends on the quality of data. Since venal data is always captured under a controlled environment, there is enough scope to obtain good quality data for veins. Parameters required for getting good quality of data are carefully studied in making the setup and are given below:

- The distance between the camera and the lamp.

- The position of the lamp.

- The fixed area for the placement of the hand.

- The orientation of the lamp once clenched within the palm.

- The focal length and the exposure time of the camera lens.

In our experiment, a simple digital SLR camera combined with an infra-red filter has been used for data acquisition. Also it makes use of a low cost night vision lamp of wavelength 940 nm. The proposed set-up is a modest wooden box with a hollow rod lodged in the middle accommodating the infra-red lamp. The camera is fixed at a perpendicular to the light source pre-adjusted to a fixed height of 10 inches above the light source. The camera is held on a tripod attached to the box. The robustness and the flat face of the night vision lamp provides for a sturdy plinth for the subject's hand. The sensor here is kept on the opposite side of the light source as shown in Fig. 2. This design has specific advantages. The subject has to place his palm on the plinth surface, to provide image. If the camera is not able to pick up the pattern the attendant can immediately rectify the hand position. The image can be captured only when the camera can pick up the veins.

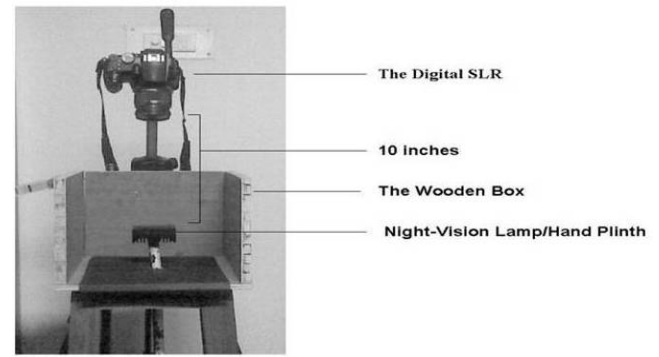

Figure 2. Experimental Setup

The setup prepared for the proposed system is not only cost effective but also meets the requirement for good quality data acquisition. It is found that through this camera along with the mentioned light the veins appear black. The light source is placed behind the surface to be captured. This helps to make an ideal data scheme and standard, as all parameters can be fixed. Unlike [8], [11], [12] and [14] where infra-red light has been reflected or transmitted to the desired surface, this paper proposes an absorption-based technique to acquire images. The proposed technique provides a good quality image regardless of the skin color or, any aberrations or discolorations, on the surface of the hand. In this technique the veins pop out when the hand is fisted and it becomes much easier to capture high contrast images. The time and cost of image processing therefore can be kept to a minimum. Since the veins are illuminated from the behind and captured from the other side, any anomalies in the skin of the palm (including the natural palm lines) would not interfere in the pattern. The image capturing however would be limited by anomalies on the palma dorsa itself, like tattoos etc. On the other hand, skin color or the gradual change of it (due to diseases or sunlight, etc.) or the gain and loss of weight would not hamper the pattern collection process.

Since the light illuminates the entire hand, it is a common notion that the veins in the front of the hand might interfere with the pattern at the back. However, it is crucial to note, that infra-red light does not make the hand transparent. It simply illuminates the hemoglobin in the veins, which appear black. The partition of the bone between the two planes in the front and the back of the hand, does not allow interference. And since the sensor is always facing the dorsal surface, it is the only surface to be captured.

The only factor due to which an inconsistency can occur during image acquisition is the size of a subject's hand since there is no control over the size and thickness of a subject's hand in a practical scenario. Therefore, the exact distance between the object and the camera's lens can never be pre-determined or fixed. To handle this situation, a necessary arrangement has been made in the setup. The focal shift of the camera which can be fine tuned to the order of millimeters ensures the *relative* prevalence of the desired conditions.

### B. Data Pre-Processing

The color image acquired through a camera generally contains some additional information which is not required to obtain the venal pattern. So there is a need to extract the region of interest from the acquired image and finally to convert into a noise free thinned image from which one can generate the venal tree. Badawi [1] has considered the skin component of the image as the region of interest (ROI). Wang and Leedham [11] used anthropometric points of a hand to segregate an ROI from the acquired images. Most similar works based on ROI selection employ arbitrary and inconsistent techniques and so end up enhancing manual intervention during processing [8]. This extracted region is used for further processing to obtain the venal pattern. This section presents the method followed to extract the ROI and then to obtain the venal tree. It consists of four major tasks and they are (i) Segmentation (ii) Image Enhancement and Binarization (iii) Dilation and Skeletonization and (iv) Venal pattern generation.

The segmentation technique used in this paper to segregate the skin area from the acquired image selects the ROI in a systematic manner and it also, successfully gets rid of all manual intervention. It fuses a traditional technique based on active contouring [5] with a common cropping technique. It works on the principle of intensity gradient, where the user initializes a contour around the object, for it to detect the boundary of the object easily. A traditional active contour is defined as a parametric curve $v(s) = [x(s), y(s)], s \in [0, 1]$, which minimizes the following energy functional.

$$E_{contour} = \int_0^1 \frac{1}{2}(\eta_1 |v'(s)|^2 + \eta_2 |v''(s)|^2) + E_{ext}(v(s)) ds \quad (1)$$

where $\eta_1$ and $\eta_2$ are weighing constants to control the relative importance of the elastic and bending ability of the active contour respectively; $v'(s)$ and $v''(s)$ are the first and second order derivatives of $v(s)$, and $E_{ext}$ is derived from the image so that it takes smaller values at the feature of interest that is edges, object boundaries etc. For an image $I(x, y)$, where $(x, y)$ are spatial co-ordinates, typical external energy can be defined as follows to lead the contour towards edges:

$$E_{ext} = -|\nabla I(x, y)|^2 \quad (2)$$

where $\nabla$ is gradient operator. For color images, we estimate the intensity gradient which takes the maximum of the gradients of *R, G and B* bands at every pixel, using:

$$\nabla I = \max(|\nabla R|, |\nabla G|, |\nabla B|) \quad (3)$$

The gradient obtained using the above equation gives better edge information. An active contour that minimizes $E_{contour}$ must satisfy the following Euler equation:

$$\eta_1 v''(s) - \eta_2 v^{iv}(s) - \nabla E_{ext} = 0 \quad (4)$$

where $v''(s)$ and $v'''(s)$ are the second and fourth order derivatives of $v(s)$. The above equation can also be viewed as a force balancing equation, $F_{int} + F_{ext} = 0$ where,

$$F_{int} = \eta_1 v''(s) - \eta_2 v^{iv}(s) \quad (5)$$

and

$$F_{ext} = -\nabla E_{ext} \quad (6)$$

$F_{int}$, the internal force is responsible for the stretching and bending and $F_{ext}$, the external force, attracts the contour towards the desired features in the image. The active contour deforms itself with time to exactly fit around the object. It can thus be represented as a time varying curve.

$$v(s, t) = [x(s, t), y(s, t)] \quad (7)$$

where $s \in [0, 1]$ is arc length and $t \in \mathbb{R}^+$ is time.

Active contouring helps the contours to settle at the object boundary. It is then followed by the iterative use of a cropping tool which helps extract the object automatically and al most flawlessly (3). It can be noted, that the active contouring snake has been modified from its regular run. Instead of initiating the snake from the outside in, it is run in reverse, after initiating it from the centre of the image. This initiation is always done automatically.

The extracted ROI of the colored image is converted into a grey scale image by the technique given in [3] as shown in Fig. 3.The segmented grey scale has been enhanced using Gaussian filtering technique and is then normalized by converting it to an image having a pre defined mean and variance. The resultant image is then binarized by mean filtering. However, it may contain noises like salt and pepper, blobs or stains, etc. Median filtering is used to remove salt and pepper type noises. Eventually a grey scale image has been denoised and binarized as given in Fig. 4.

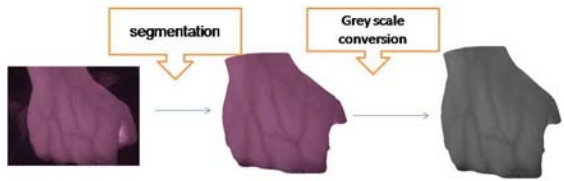

Figure 3. Automatically Segmented Image

The image may consist of a few edges of the vein pattern that may have been falsely eroded during filtering. These edges are reconnected by dilation, i.e., running a disk of ascertained radius over the obtained pattern. Then these obtained images are skeletonized. Each vein is reduced to its central pixel and their thickness is reduced to 1 pixel size only. A skeletonized image can hence, be obtained (see Fig. 5).

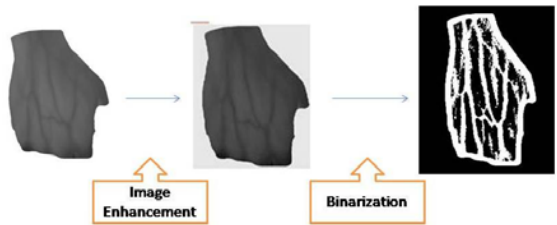

Figure 4. Enhanced and Binarized Grey Scale Image

In order to obtain only desired components amongst veins, all connected components are labeled and others are discarded. The CCL (Connected Component Labeling) algorithm [6] is modified to determine all the connected components in an image. This modified algorithm detects and removes all isolated and disconnected components of size less than a specified threshold.

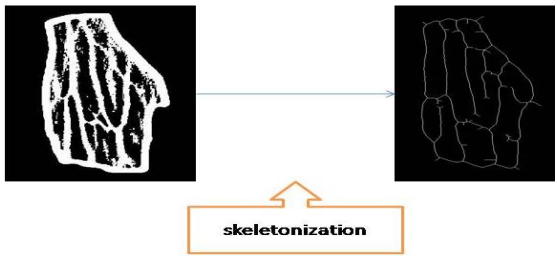

Figure 5. The Binarized image can be processed to give the vein skeleton in the hand

From the skeleton of the hand, the skeletonized veins are extracted. A vertical and a horizontal line (one pixel thick) are run through each coordinate of each image alternatively. The coordinates of the first and the last image pixels encountered by the line, in both axes, are stored. These coordinates were later turned black and the venal tree was extracted. The modified connected component labeling (CCL) algorithm is executed again to remove all disconnected isolated components from the final skeleton.

### C. Feature Extraction

This section presents a technique which extracts the forkings from the skeleton image by examining the local neighborhood of each ridge pixel using a 3X3 window. It can be seen from the preprocessing image that an ROI contains some thinned lines/ridges. These ridges representing vein patterns can be used to extract features. Features like ridge forkings are determined by computing the number of *arms originating from a pixel*. This can be represented as *A*. The *A* for a pixel *P* can be given as:

$$A = 0.5 \sum_{i=1}^{8} |P_i - P_{i+1}|, P_9 = P_1$$

(8)

For a pixel P, its eight neighboring pixels are scanned in an anti-clockwise direction as follows:

| $P_4$ | $P_3$ | $P_2$ |
|-------|-------|-------|
| $P_5$ | P     | $P_1$ |
| $P_6$ | $P_7$ | $P_8$ |

A given pixel P is termed as a ridge forking for a vein pattern if the value of A for the pixel is 3 or more. This ridge forking pixel is considered as a feature point which can be defined by (x, y, θ) where x and y are coordinates and θ is the orientation with respect to a reference point.

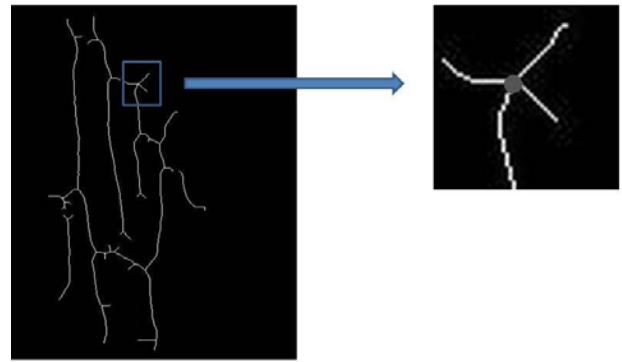

Figure 6. Four Arms emitting from a forking point

The proposed method for calculating A can accommodate three or four or more arms emitting out of a forking point. Cases where four arms emit from a forking point are common, as shown in Fig. 6. Fig.7 shows the final image of the extracted vein pattern with all forking points marked.

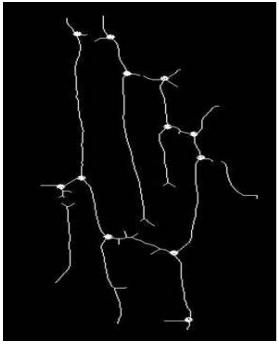

Figure 7. The Final Vein Pattern with all Forking Points Marked

### D. Matching Strategy

Suppose, *N* and *M* are two patterns having *n* and *m* features respectively. Then the sets *N* and *M* are given by:

$$N = \{(x_1, y_1, \theta_1), (x_2, y_2, \theta_2), (x_3, y_3, \theta_3), \ldots, (x_n, y_n, \theta_n)\}$$

$$M = \{(a_1, b_1, \varphi_1), (a_2, b_2, \varphi_2), (a_3, b_3, \varphi_3), \ldots, (a_m, b_m, \varphi_m)\}$$

where $(x_i, y_i, \theta_i)$ and $(a_j, b_j, \varphi_j)$ are the corresponding features in pattern *N* and *M* respectively. For a given minutiae $(x_i, y_i, \theta_i)$ in *N*, it first determines a minutiae $(a_j, b_j, \varphi_j)$ such that the distance $\left| \sqrt{(x_i - a_j)^2 + (y_i - b_j)^2} \right|$ is minimum for all *j*, $j=1,2,3 \ldots, m$. Let the distance be $sd_i$ and the corresponding difference between two directions be $dd_i$, where $dd_i = \left| \theta_i - \varphi_j \right|$.

This is done for all features in *N*. To avoid the selection of same feature in *M* for a given minutiae in *N*, one can follow the following procedure. Suppose, for the $i^{th}$ feature in *N*, one gets $sd_i$ for the $j^{th}$ feature in *M*. Then, in order to determine $sd_{i+1}$, one considers all features in *M* which are not selected in $sd_1$, $sd_2 \ldots sd_i$. Let *A* be a binary array of *n* elements satisfying

$$A[i] = \begin{cases} 1 & if\ (sdi \le ti)\ and\ (ddi \le t2) \\ 0 & otherwise \end{cases}$$

where $t_1$ and $t_2$ are predefined thresholds. The threshold values defined by $t_1$ and $t_2$ are necessary to compensate for the unavoidable errors made by feature extraction algorithms and to account for the small plastic distortions that cause the minutiae positions to change. These are thresholds determined by averaging the different feature shifts based on intensive testing.

Then the percentage of match is obtained for the pattern *N* having *n* features against the pattern can be computed by

$$V = \frac{\sum_{i=1}^{n} A[i]}{n} X\ 100\ \%$$

(9)

If *V* is more than a given threshold then one can draw the conclusion that both the patterns are matched.

### III. EXPERIMENTAL RESULTS

The proposed system has been tested against the IITK database to analyze its performance. The database consists of 1750 images obtained from 341 individuals under controlled environment. Out of these 1750 images, 341 are used as query samples. A graph is plotted for the achieved accuracy against the various threshold values as shown in Fig.8. It is observed that the maximum accuracy of 99.26% can be achieved at the threshold value, *T*, of 25. Graphically, it is also found in Fig. 9 that the value of FRR for which the system achieves maximum accuracy is 0.03%. Finally, the ROC curve is taken between the values of GAR and the FAR is given in Fig. 10.

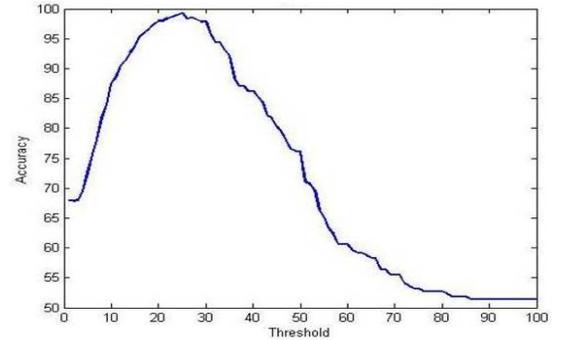

Figure 8. Graph Accuracy

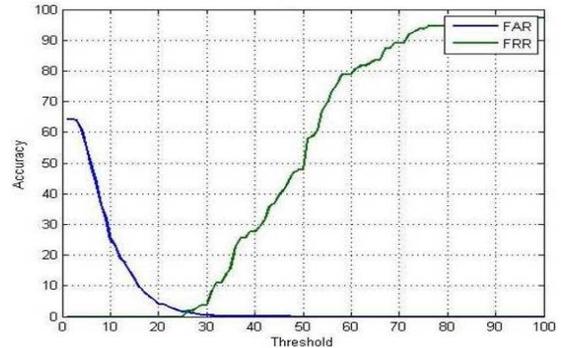

Figure 9. Graph indicating FAR and FRR

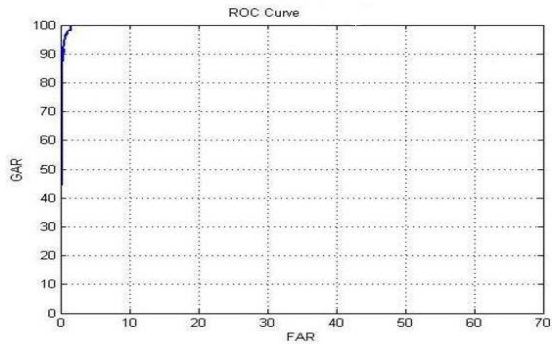

Figure 10. The ROC Curve- GAR v/s FAR

## IV. CONCLUSION

This paper has proposed a new absorption based vein pattern recognition system. It has a very low cost data acquisition set up, compared to that used by others. The system has made an attempt to handle issues such as effects of rotation and translation on acquired images, minimizing the manual intervention to decide on the verification of an individual. It has been tested in a controlled environment and against a dataset of 1750 samples obtained from 341 subjects. The experimental results provide an excellent accuracy of 99.26% with FRR 0.03%. This is found to be comparable to most previous works [2] [11] [12] [13] and is achieved through a technique which is found to be much simpler.

### AUTHORS PROFILE

**Mohit Soni** graduated from the Delhi University with an honors degree in Botany and Biotechnology. He received his Masters degree in Forensic Science from the National Institute of Criminology and Forensic Sciences, New Delhi. Thereafter he received a research fellowship from the Directorate of Forensic Sciences, New Delhi in 2006 and is currently pursuing his Doctoral degree in Biometrics and Computer Science from the Uttar Pradesh Technical University, Lucknow.

**Sandesh Gupta** received his Bachelors in Technology from the University Institute of Engineering and Technology, C.S.J.M University Kanpur in 2001. He is working currently as a lecturer for the computer science department in the same institution and is pursuing his PhD from the Uttar Pradesh Technical University, Lucknow.

**M S Rao** is a well known forensic scientist of the country and started his career in Forensic Science in the year 1975 from Orissa Forensic Science Laboratory. He carried extensive R&D work on Proton Induced X-Ray Emission (PIXE) in Forensic Applications during 1978-1981. He was appointed as Chief Forensic Scientist to the Government of India in 2001. He was Secretary and Treasurer for the Indian Academy of Forensic Sciences from 1988 to 2000 and is now the President of the Academy. He was convener of the Forum on Forensic Science of the Indian Science Congress during 1992 and 2001. He is the Chairman of the Experts Committee on Forensic Science.

**Phalguni Gupta** received the Doctoral degree from Indian Institute of Technology Kharagpur, India in 1986. Currently he is a Professor in the Department of Computer Science & Engineering, Indian Institute of Technology Kanpur (IITK), Kanpur, India. He works in the field of biometrics, data structures, sequential algorithms, parallel algorithms, on-line algorithms. He is an author of 2 books and 10 book chapters. He has published more than 200 papers in International Journals and International Conferences. He is responsible for several research projects in the area of Biometric Systems, Image Processing, Graph Theory and Network Flow.